\documentclass[11pt]{article}
\usepackage{tabularx}
\usepackage{tabularx}
\usepackage{booktabs}
\usepackage{amsmath}
\usepackage{graphicx, makecell, booktabs, multirow}
\usepackage{array}
\usepackage{float}
\usepackage[final]{acl}
\usepackage{amsmath}
\usepackage{amsmath,amssymb,amsthm}
\usepackage{times}
\usepackage{latexsym}

\usepackage[T1]{fontenc}

\usepackage[utf8]{inputenc}

\usepackage{microtype}

\usepackage{inconsolata}

\usepackage{graphicx}

\title{CLM-Bench: Benchmarking and Analyzing Cross-lingual Misalignment of LLMs in Knowledge Editing}

\author{
Yucheng Hu\textsuperscript{1} \quad
Wei Zhou\textsuperscript{1} \quad
Juesi Xiao\textsuperscript{2} \\[1ex]
\textsuperscript{1}Tianjin University, School of Future Technology \\ 
\textsuperscript{2}Tianjin University, College of Intelligence and Computing \\[1ex]
\texttt{yc\_h666@tju.edu.cn}
}

\begin{document}

\maketitle
\begin{abstract}
Knowledge Editing (KE) has emerged as a promising paradigm for updating facts in Large Language Models (LLMs) without retraining. However, progress in Multilingual Knowledge Editing (MKE) is currently hindered by biased evaluation frameworks. We observe that existing MKE benchmarks are typically constructed by mechanically translating English-centric datasets into target languages (e.g., English-to-Chinese). This approach introduces translation artifacts and neglects culturally specific entities native to the target language, failing to reflect the true knowledge distribution of LLMs. To address this, we propose CLM-Bench, a culture-aware benchmark constructed using a native Chinese-first methodology. Unlike previous works, we curate 1,010 high-quality CounterFact pairs rooted in Chinese cultural contexts (covering domains like history and literature) and subsequently align them with English counterparts. Using CLM-Bench, we conduct extensive experiments on representative LLMs (e.g., Llama-3, Qwen2) and reveal a significant Cross-lingual Misalignment: edits in one language function independently and fail to propagate to the other. We further provide a geometric explanation via layer-wise representation analysis, demonstrating that edit vectors for Chinese and English are nearly orthogonal—residing in disjoint subspaces—while mixed-lingual editing exhibits perfect linear additivity of these vectors. Our findings challenge the effectiveness of current methods in cross-lingual transfer and underscore the importance of culturally native benchmarks.\footnote{Data are available at \url{https://github.com/kekeaii/CLM_bench}}
\end{abstract}

\section{Introduction}

Large Language Models (LLMs) have demonstrated remarkable abilities in memorizing and utilizing world knowledge \cite{chang2024survey,ouyang2022training}.
However, a fundamental limitation persists: the parametric knowledge within these models is static. 
As the world evolves, the facts stored in LLMs become outdated or incorrect \cite{chen2024robust,wang2024factuality,tang2025evowiki}.
To address this, Knowledge Editing (KE) has emerged as a promising paradigm, with the aim of precisely modifying specific facts in a model without the prohibitive cost of retraining or the degradation of catastrophic forgetting \cite{gupta2024model,wang2024easyedit}.

\begin{figure}[t]
    \centering
    \includegraphics[width=0.8\linewidth]{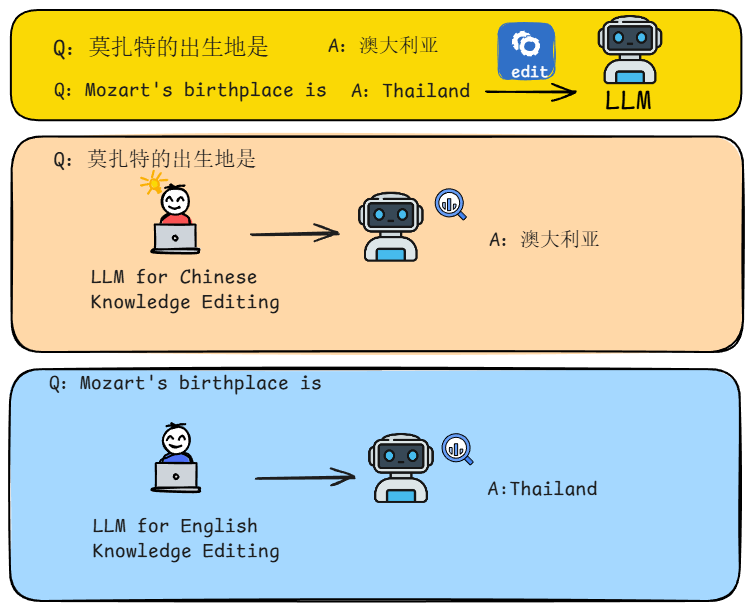}
    \caption{This figure illustrates Chinese–English editing independence: Applying the same edit with different target values yields different Chinese and English outputs.}
    \label{fig:misalignment}
\end{figure}

Although current KE methods, such as ROME \cite{meng2022locating} and Wise \cite{wang2024wise}, have achieved high reliability in monolingual settings (primarily English), they largely overlook a critical dimension of reality in multilinguality.
As LLMs are increasingly deployed in cross-lingual scenarios \cite{wang2024probing,doddapaneni2025cross}, a successful edit should ideally propagate across languages. 
For instance, updating the ``Prime Minister of the UK" in English should spontaneously correct the corresponding knowledge when queried in Chinese. 
However, recent studies in Multilingual Knowledge Editing (MKE) suggest that cross-lingual transfer is far from trivial \cite{wang2024cross,raheja2024medit}.

We identify several limitations in current multilingual knowledge editing research 
that motivate our work. First, existing benchmarks are mainly based on translated 
datasets. Most MKE datasets, such as ZsRE \cite{wang2024cross}, are constructed by 
translating English datasets into other languages, introducing 
"translationese artifacts" and disconnecting from culturally native entities 
(e.g., Chinese idioms or local celebrities) \cite{liu2025translation}. 
Second, and more critically, the underlying 
mechanism of cross-lingual misalignment remains unexplained. While previous work 
observes that editing one language often fails to transfer to another, the geometric 
relationship between language-specific representations has not been rigorously 
analyzed \cite{kargaran2025mexa}.

To bridge these gaps, we propose \textbf{CLM-Bench} (Cross-Lingual Misalignment Benchmark), a comprehensive framework designed to evaluate and analyze the alignment of knowledge editing across languages. 
Unlike previous works, we construct a native Chinese-first CounterFact dataset consisting of 1,010 high-quality pairs across diverse domains (History, Literature, Science, etc.), which are then aligned with English counterparts.
This ensures that the evaluation reflects genuine native language understanding rather than translation artifacts.
Using CLM-Bench, we conduct extensive experiments on representative LLMs (e.g., Llama-3, Qwen2) using state-of-the-art editing methods.
Our empirical results reveal a significant phenomenon of Cross-lingual misalignment as shown in Figure~\ref{fig:misalignment}: 
the independence and linear additivity of Chinese and English edits

Further, we provide a geometric explanation for this misalignment via layer-wise representation analysis. 
By probing the hidden states at deeper layers, we discover two critical geometric properties of editing vectors: (1) The edit vectors for Chinese and English representations are nearly orthogonal.
This explains why cross-lingual transfer fails—the optimization direction for one language has almost no projection onto the other.
(2) Moreover, the resulting vector is an almost perfect linear sum of the individual language vectors when we perform mixed-language editing.
This insight suggests that multilingual knowledge in current LLMs is stored in disjoint subspaces that do not naturally interact during gradient-based or locate-and-edit updates, challenging the ``interlingua" hypothesis in the context of parameter editing.
Overall, our contributions are as follows:

\begin{itemize}
    \item We introduce CLM-Bench, a culture-aware and counterfact benchmark for evaluating cross-lingual knowledge editing, overcoming the limitations of translation-based datasets.

    \item We provide a systematic evaluation of current editing methods, uncovering the asymmetry and independence of editing effects between Chinese and English.

    \item We offer a novel geometric interpretation based on vector orthogonality and linearity, providing an experimental foundation for why current methods fail in cross-lingual transfer and how mixed-editing strategies succeed.
\end{itemize}

\begin{figure*}[t]
    \centering
    \includegraphics[width=0.9\linewidth]{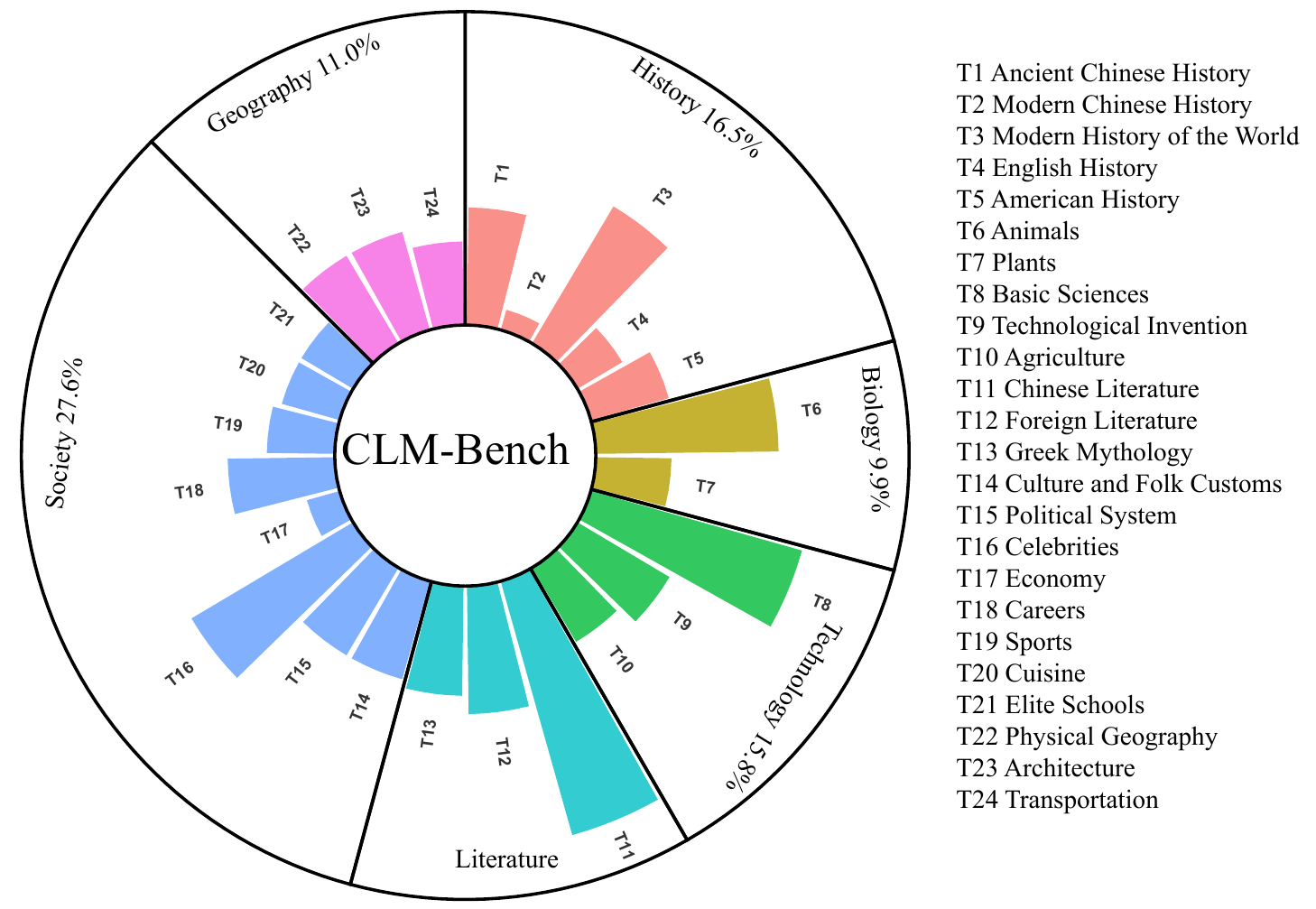}
    \caption{Overview of CLM-Bench, covering 6 subdomains and 24 domains.}
    \label{fig:bench}
\end{figure*}

\section{Related Work}

\subsection{Knowledge Editing Methods}

Knowledge editing approaches include:  
\textbf{Meta-learning methods} such as MEND \cite{mitchell2022fastmodeleditingscale}, which learnEACL editing rules from data;  
\textbf{Memory-augmented methods} such as SERAC \cite{mitchell2022memorybasedmodeleditingscale}, which store edits externally;  
\textbf{Locate-and-edit methods}, now the most effective for persistent edits.
Among locate-and-edit approaches:  
\textbf{ROME} \cite{meng2022locating} performs rank-one updates to targeted MLP layers;  
\textbf{MEMIT} \cite{meng2022memit} extends this to multi-fact batch editing;  
\textbf{AlphaEdit} \cite{fang2024alphaedit} introduces a null-space constraint to reduce interference;  
\textbf{PMET} \cite{li2024pmetprecisemodelediting} refines Transformer parameter transformations for more precise edits.

While effective in monolingual (mainly English) settings, their cross-lingual behavior remains understudied\cite{durrani2025editinglanguagessurveymultilingual}.

\subsection{Knowledge Editing Datasets}

Most existing knowledge editing benchmarks are monolingual and predominantly English. 
CounterFact \cite{meng2022locating} and ZsRE \cite{levy2017zeroshotrelationextractionreading} are the most widely used datasets, providing factual triples and counterfactual queries for evaluating locality, specificity, and generalization of edits. 
To move beyond English, recent work has introduced multilingual variants. 
Bi-ZsRE \cite{wang2024cross} extends ZsRE to Chinese--English pairs, enabling bilingual consistency evaluation. 
\citet{beniwal-etal-2024-cross} further investigate multilingual editing across diverse scripts (e.g., Latin, Indic) using mBERT and XLM-RoBERTa, highlighting cross-lingual challenges not captured by monolingual benchmarks.

However, translation-based datasets exhibit known limitations. For multilingual NLP benchmarks, studies such as \cite{liu2025translation} show that translation may introduce “translationese” artifacts and distort cultural/language nuances, reducing linguistic naturalness. They also rarely include language-specific entities or relations that stress-test multilingual factual grounding. Consequently, such limitations may obscure cross-lingual failure modes in knowledge-editing evaluations and motivate the development of culturally grounded multilingual benchmarks.

\section{CLM-Bench}

\subsection{Data Construction Pipeline}
The construction of CLM-Bench mainly consists of three components: Data Source, Generation Methodology, and Post-processing.More details can be found in Appendix~\ref{app:data}.
\subsubsection{Data Sources}
The data sources for this dataset are primarily divided into two categories: (a) \textbf{LLM-assisted Generation.} We utilized the Deepseek-R1 large language model as the core tool to generate the original content. (b) \textbf{Open-source Datasets.} Our research incorporated partial data from the open-source dataset ZsRE\cite{levy2017zeroshotrelationextractionreading} and CounterFact\cite{meng2022locating}

\subsubsection{Generation Methodology}
Our generation methodology consists of the following steps: (a) \textbf{Data Design.} We designed a structured data format and established specific linguistic guidelines for each core field. Each entry contains the following core fields: prompt, target\_new, ground\_truth, subject, and rephrase\_prompt. (b) \textbf{Knowledge Categorization.} To ensure the generated data is not concentrated in a single domain, we categorized the intended data into 24 distinct knowledge domains and crafted specific prompts for the LLM accordingly. (c) \textbf{Model Generation.} We used the Deepseek-R1 LLM to assist in data generation. Initial content and its corresponding English version were produced using manually designed prompts. Approximately 1,100 raw data entries were generated through multi-turn dialogues. (d) \textbf{Merging.} We extracted the loc and loc\_ans sections from the open-source dataset and merged them with our generated data to complete the construction of the full dataset.

\subsubsection{Post-processing}
To ensure data quality, we implemented the following post-processing steps: (a) \textbf{Deduplication.} Manual deduplication was performed on the approximately 1,100 entries, removing samples with duplicate prompt and subject fields. (b) \textbf{Quality Review.} Reviewers ensured the ground\_truth was accurate, the target\_new was an incorrect answer forming a clear contrast with the ground truth. Simultaneously, the English versions of the data were manually reviewed to guarantee grammatical correctness, linguistic fluency, naturalness, and compliance with format requirements. (c) \textbf{Standardization.} All data was uniformly formatted using the JSON standard.

\subsection{Dataset Statistics and Quality}
Figure \ref{fig:bench} presents the statistical overview of our dataset. The final dataset comprises 1,010 high-quality entries, covering 24 different knowledge domains to ensure comprehensive evaluation. Chinese Literature constitutes the largest portion at 10.1\%, while Modern Chinese History accounts for the smallest portion at 0.8\%.

Our dataset holds advantages over traditional translated datasets in terms of entity coverage and language naturalness: (a) \textbf{Entity Coverage.} Traditional translated datasets are often built by directly translating an English benchmark into the target language. Consequently, the entities they contain exhibit a strong bias towards English. For instance, they may include numerous obscure foreign personal and place names while severely lacking entities important in Chinese. Our dataset carefully selects entities significant in both Chinese and English contexts for data construction, ensuring balanced entity coverage. (b) \textbf{Language Naturalness.} Traditional translated datasets inevitably suffer from translation errors, leading to rigid, unnatural expressions that poorly reflect the authentic usage patterns of native speakers. Our dataset employs Deepseek-R1 for initial translation, followed by manual review and editing, ensuring fluency and naturalness in both languages.

\begin{table*}[th]
\centering
\caption{Model Editing Evaluation Results.The results show that Chinese editing accuracy is approximately equal to mixed editing accuracy in Chinese, and English editing accuracy is approximately equal to mixed editing accuracy in English.}
\label{tab:data1}
\scriptsize
\renewcommand{\arraystretch}{1.0}
\setlength{\tabcolsep}{3pt}
\begin{tabular}{@{} l l l *{9}{c} @{}}
\toprule
\multirow{2}{*}{Model} & 
\multirow{2}{*}{Methods} & 
\multirow{2}{*}{\makecell{Editing\\Language}} & 
\multirow{2}{*}{\makecell{zh\\score}} & 
\multirow{2}{*}{\makecell{en\\score}} & 
\multicolumn{3}{c}{zh} & \multicolumn{3}{c}{en} & 
\multirow{2}{*}{trans} \\
\cmidrule(lr){6-8} \cmidrule(l){9-11}
 & & & & & 
\makecell{Reliability} & 
\makecell{Generality} & 
\makecell{Locality} & 
\makecell{Reliability} & 
\makecell{Generality} & 
\makecell{Locality} & \\
\midrule
\multirow{9}{*}{Llama3-8B} & 
\multirow{3}{*}{MEMIT} & Chinese & 61.71\% & 43.44\% & \textbf{63.21\%} & 56.74\% & 65.17\% & 19.55\% & 18.69\% & 92.09\% & 30.93\% \\
 & & English & 59.33\% & 68.31\% & 44.42\% & 44.91\% & 88.67\% & \textbf{75.47\%} & 39.08\% & 90.37\% & 58.86\% \\
 & & mix & 61.24\% & 66.83\% & \textbf{63.21\%} & 57.07\% & 63.43\% & \textbf{72.73\%} & 38.89\% & 88.86\% & 86.91\% \\
\cmidrule(lr){2-12}
 & \multirow{3}{*}{AlphaEdit} & Chinese & 1.84\% & 4.04\% & \textbf{1.85\%} & 1.49\% & 2.19\% & 4.12\% & 3.97\% & 4.03\% & 44.90\% \\
 & & English & 15.96\% & 38.00\% & 14.79\% & 13.90\% & 19.20\% & \textbf{64.98\%} & 36.53\% & 12.50\% & 22.76\% \\
 & & mix & 1.54\% & 11.38\% & \textbf{1.05\%} & 1.36\% & 2.20\% & \textbf{19.58\%} & 9.24\% & 5.33\% & 5.36\% \\
\cmidrule(lr){2-12}
 & \multirow{3}{*}{PMET} & Chinese & 51.17\% & 44.35\% & \textbf{41.83\%} & 41.46\% & 70.23\% & 19.50\% & 18.09\% & 95.45\% & 46.62\% \\
 & & English & 58.48\% & 42.54\% & 43.15\% & 43.14\% & 89.16\% & \textbf{19.60\%} & 18.99\% & 89.03\% & 45.42\% \\
 & & mix & 53.18\% & 42.36\% & \textbf{43.18\%} & 42.92\% & 73.45\% & \textbf{19.94\%} & 19.54\% & 87.61\% & 46.18\% \\
\midrule
\multirow{9}{*}{Mistral-7B} & 
\multirow{3}{*}{MEMIT} & Chinese & 8.62\% & 46.76\% & \textbf{6.99\%} & 6.69\% & 12.18\% & 27.35\% & 27.36\% & 85.56\% & 25.56\% \\
 & & English & 40.56\% & 58.96\% & 33.85\% & 32.71\% & 55.12\% & \textbf{57.74\%} & 45.13\% & 74.02\% & 58.62\% \\
 & & mix & 7.57\% & 55.81\% & \textbf{6.73\%} & 5.99\% & 9.99\% & \textbf{55.39\%} & 43.43\% & 68.60\% & 12.15\% \\
\cmidrule(lr){2-12}
 & \multirow{3}{*}{AlphaEdit} & Chinese & 14.25\% & 44.25\% & \textbf{14.30\%} & 13.18\% & 15.26\% & 26.76\% & 25.99\% & 80.01\% & 53.44\% \\
 & & English & 34.96\% & 41.12\% & 33.96\% & 33.64\% & 37.27\% & \textbf{43.58\%} & 35.65\% & 44.13\% & 77.93\% \\
 & & mix & 11.53\% & 40.64\% & \textbf{13.00\%} & 11.33\% & 10.26\% & \textbf{43.29\%} & 35.72\% & 42.90\% & 30.03\% \\
\cmidrule(lr){2-12}
 & \multirow{3}{*}{PMET} & Chinese & 8.96\% & 2.83\% & \textbf{13.02\%} & 12.91\% & 0.96\% & 1.32\% & 0.93\% & 6.23\% & 10.14\% \\
 & & English & 7.04\% & 16.09\% & 5.73\% & 5.48\% & 9.92\% & \textbf{21.01\%} & 15.16\% & 12.09\% & 27.27\% \\
 & & mix & 10.70\% & 15.16\% & \textbf{15.53\%} & 15.74\% & 0.83\% & \textbf{20.95\%} &  13.10\% & 11.43\% & 74.13\% \\
\midrule
\multirow{9}{*}{Qwen2-7B} & 
\multirow{3}{*}{MEMIT} & Chinese & 58.56\% & 38.46\% & \textbf{67.56\%} & 59.81\% & 48.32\% & 22.36\% & 21.21\% & 71.82\% & 33.10\% \\
 & & English & 52.43\% & 69.29\% & 46.60\% & 45.61\% & 65.08\% & \textbf{80.69\%} & 57.72\% & 69.45\% & 57.75\% \\
 & & mix & 55.08\% & 67.38\% & \textbf{63.92\%} & 57.13\% & 44.20\% & \textbf{82.13\%} & 58.56\% & 61.45\% & 77.83\% \\
\cmidrule(lr){2-12}
 & \multirow{3}{*}{AlphaEdit} & Chinese & 14.87\% & 17.84\% & \textbf{16.91\%} & 13.94\% & 13.75\% & 17.02\% & 16.39\% & 20.10\% & 99.35\% \\
 & & English & 38.53\% & 57.51\% & 38.31\% & 37.46\% & 38.82\% & \textbf{75.77\%} & 58.99\% & 37.77\% & 50.56\% \\
 & & mix & 8.83\% & 35.56\% & \textbf{10.23\%} & 7.69\% & 8.57\% & \textbf{54.08\%} & 40.27\% & 12.32\% & 18.92\% \\
\cmidrule(lr){2-12}
 & \multirow{3}{*}{PMET} & Chinese & 0.77\% & 15.92\% & \textbf{0.07\%} & 0.03\% & 2.20\% & 14.98\% & 12.42\% & 20.37\% & 0.47\% \\
 & & English & 0.50\% & 1.02\% & 0.27\% & 0.15\% & 1.08\% & \textbf{0.48\%} & 0.32\% & 2.27\% & 56.25\% \\
 & & mix & 0.52\% & 0.73\% & \textbf{0.17\%} & 0.40\% & 1.00\% & \textbf{0.21\%} & 0.31\% & 1.66\% & 80.95\% \\
\midrule
\multirow{9}{*}{llama2-chinese-7B} & 
\multirow{3}{*}{MEMIT} & Chinese & 13.84\% & 47.92\% & \textbf{13.23\%} & 12.07\% & 16.21\% & 30.27\% & 30.19\% & 83.31\% & 43.71\% \\
 & & English & 50.05\% & 68.72\% & 39.79\% & 39.41\% & 70.96\% & \textbf{71.42\%} & 59.41\% & 75.32\% & 55.71\% \\
 & & mix & 14.05\% & 65.29\% & \textbf{14.06\%} & 12.81\% & 15.28\% & \textbf{69.97\%} & 57.41\% & 68.50\% & 20.09\% \\
\cmidrule(lr){2-12}
 & \multirow{3}{*}{AlphaEdit} & Chinese & 8.64\% & 25.85\% & \textbf{9.54\%} & 8.60\% & 7.78\% & 19.15\% & 18.80\% & 39.60\% & 49.82\% \\
 & & English & 21.46\% & 29.04\% & 15.31\% & 15.30\% & 33.77\% & \textbf{34.07\%} & 27.53\% & 25.52\% & 44.94\% \\
 & & mix & 9.45\% & 20.01\% & \textbf{11.52\%} & 10.50\% & 6.33\% & \textbf{26.27\%} & 19.23\% & 14.52\% & 43.85\% \\
\cmidrule(lr){2-12}
 & \multirow{3}{*}{PMET} & Chinese & 50.76\% & 51.50\% & \textbf{37.64\%} & 37.25\% & 77.40\% & 29.46\% & 29.39\% & 95.66\% & 78.27\% \\
 & & English & 60.22\% & 51.39\% & 43.06\% & 42.66\% & 94.93\% & \textbf{30.00\%} & 29.85\% & 94.33\% & 69.67\% \\
 & & mix & 51.38\% & 50.88\% & \textbf{38.13\%} & 37.76\% & 78.24\% & \textbf{30.22\%} & 29.82\% & 92.60\% & 79.26\% \\
\bottomrule
\end{tabular}
\end{table*}

\section{Experimental details}
\subsection{Experimental Settings}

We evaluate cross-lingual and mixed-lingual knowledge editing across four large language models with distinct linguistic training profiles: two with substantial Chinese corpora—Llama2-7B Chinese, which was incrementally pre-trained on large-scale Chinese data to improve foundational semantic understanding~\citep{touvron2023llama2openfoundation}, and Qwen2-7B~\citep{yang2024qwen2technicalreport}—and two primarily English-dominant models, Llama3-8B~\citep{grattafiori2024llama3herdmodels} and Mistral-7B~\citep{jiang2023mistral7b}. This selection allows us to compare editing behaviors across multilingual, Chinese-oriented, and English-oriented architectures.

Because our work focuses on batch-mode model editing, where all target facts are injected into the model in a single editing pass with persistent, lifelong effects, our method primarily considers the three state-of-the-art batch editing approaches: \textbf{MEMIT}~\citep{meng2023massediting}, \textbf{AlphaEdit}~\citep{fang2024alphaedit}, and \textbf{PMET}~\cite{li2024pmetprecisemodelediting}.\textbf{MEMIT} performs large-scale batch editing by directly computing parameter updates, enabling injection of thousands of facts at once.
\textbf{AlphaEdit} adds a null-space constraint, projecting updates into the null space of preserved knowledge to reduce interference and improve stability during sequential edits.
\textbf{PMET} refines parameter transformations in Transformer models to achieve more precise and controlled batch edits.
All algorithms are implemented using the \texttt{EasyEdit} framework~\citep{wang2023easyedit}.

\begin{table*}[th]
\centering
\caption{Model Editing Evaluation Results in different batchsize}
\label{tab:data2}
\scriptsize
\renewcommand{\arraystretch}{1.0}
\setlength{\tabcolsep}{3pt}
\begin{tabular}{@{} l l l l *{9}{c} @{}}
\toprule
\multirow{2}{*}{Model} & 
\multirow{2}{*}{\makecell{Batch\\size}} & 
\multirow{2}{*}{Methods} & 
\multirow{2}{*}{\makecell{Editing\\Language}} & 
\multirow{2}{*}{\makecell{zh\\score}} & 
\multirow{2}{*}{\makecell{en\\score}} & 
\multicolumn{3}{c}{zh} & \multicolumn{3}{c}{en} & 
\multirow{2}{*}{trans} \\
\cmidrule(lr){7-9} \cmidrule(l){10-12}
 & & & & & & 
\makecell{Reliability} & 
\makecell{Generality} & 
\makecell{Locality} & 
\makecell{Reliability} & 
\makecell{Generality} & 
\makecell{Locality} & \\
\midrule
\multirow{12}{*}{Llama3-8B} & 
\multirow{3}{*}{1} & \multirow{3}{*}{MEMIT} & Chinese & 86.67\% & 44.44\% & \textbf{80.00\%} & 80.00\% & 100.00\% & 33.33\% & 0.00\% & 100.00\% & 41.66\% \\
 & & & English & 60.00\% & 66.67\% & 40.00\% & 40.00\% & 100.00\% & \textbf{100.00\%} & 0.00\% & 100.00\% & 40.00\% \\
 & & & mix & 88.89\% & 66.67\% & \textbf{100.00\%} & 100.00\% & 66.67\% & \textbf{100.00\%} & 0.00\% & 100.00\% & 100.00\% \\
\cmidrule(lr){2-13}
 & \multirow{3}{*}{10} & \multirow{3}{*}{MEMIT} & Chinese & 75.83\% & 43.45\% & \textbf{73.00\%} & 64.33\% & 90.16\% & 21.00\% & 12.67\% & 96.67\% & 28.77\% \\
 & & & English & 59.70\% & 71.23\% & 43.44\% & 38.17\% & 97.50\% & \textbf{80.00\%} & 37.67\% & 96.03\% & 54.30\% \\
 & & & mix & 89.77\% & 73.00\% & \textbf{94.17\%} & 88.83\% & 86.31\% & \textbf{87.50\%} & 31.50\% & 100.00\% & 92.92\% \\
\cmidrule(lr){2-13}
 & \multirow{3}{*}{100} & \multirow{3}{*}{MEMIT} & Chinese & 71.87\% & 43.93\% & \textbf{74.41\%} & 63.24\% & 77.96\% & 18.80\% & 16.10\% & 96.88\% & 25.27\% \\
 & & & English & 60.93\% & 71.34\% & 42.93\% & 42.53\% & 97.33\% & \textbf{82.83\%} & 36.06\% & 95.14\% & 51.83\% \\
 & & & mix & 74.50\% & 71.62\% & \textbf{77.20\%} & 66.45\% & 79.85\% & \textbf{80.25\%} & 37.32\% & 97.29\% & 96.20\% \\
\cmidrule(lr){2-13}
 & \multirow{3}{*}{1000} & \multirow{3}{*}{MEMIT} & Chinese & 61.71\% & 43.44\% & \textbf{63.21\%} & 56.74\% & 65.17\% & 19.55\% & 18.69\% & 92.09\% & 30.93\% \\
 & & & English & 59.33\% & 68.31\% & 44.42\% & 44.91\% & 88.67\% & \textbf{75.47\%} & 39.08\% & 90.37\% & 58.86\% \\
 & & & mix & 61.24\% & 66.83\% & \textbf{63.21\%} & 57.07\% & 63.43\% & \textbf{72.73\%} & 38.89\% & 88.86\% & 86.91\% \\
\bottomrule
\end{tabular}
\end{table*}

Our primary experiments use a batch of 1000 edits in each monolingual setting (Chinese-only and English-only). To further examine multilingual interactions, we introduce a \textbf{mixed-lingual editing setup} consisting of 2000 bilingual edit triplets, where each fact is injected simultaneously using both a Chinese and an English trigger. This setup probes whether editing jointly in two languages improves or compromises cross-lingual consistency. Additional ablations are performed on Llama3-8B using MEMIT to assess: 
(1) scale sensitivity with smaller batch sizes, and 
(2) layer-specific effects by editing different MLP layers. 
Similar layer ablations are also conducted on Qwen2-7B to verify consistency 
across model architectures. All evaluations are conducted in a lifelong (sequential) editing protocol, where previous edits remain active.
\subsection{Evaluation Metrics}

Following standard practice in knowledge editing
\cite{meng2022locating,meng2023massediting,muller2023crosslingual}, 
we evaluate edited models using eight metrics grouped into four
dimensions, plus a cross-lingual transfer score.

\paragraph{Basic Performance.}
\texttt{zh\_score} and \texttt{en\_score} represent the overall editing performance for Chinese and English queries, respectively. Each score is computed as the average of the three core editing criteria: \textit{locality}, \textit{reliability}, and \textit{generality}.
\paragraph{Reliability.}
Reliability measures the success rate of edits in both Chinese and English. It is computed by comparing the model's predicted tokens against the ground-truth answers and calculating the corresponding accuracy.
\paragraph{Generality.}
Generality assesses the model's ability to generalize an edit across paraphrased queries. Specifically, it evaluates whether the edited model can correctly answer rephrased versions of the original questions. The accuracy is computed using the same procedure as in reliability.
\paragraph{Locality.}
Locality quantifies the preservation of unrelated knowledge. It is defined as the accuracy on unrelated queries and is measured by the similarity between the model's responses before and after editing for the same control questions.
\paragraph{Cross-lingual Transfer.}
Cross-lingual transfer evaluates the alignment between Chinese and English editing performance. It is measured by comparing the accuracy of edits applied in Chinese with the accuracy of the corresponding edits applied in English.

\section{Experimental Results}

\subsection{The Misalignment Phenomenon}

Our experimental results confirm the misalignment between knowledge editing success and multilingual generation capabilities previously observed in BiZsre\cite{wang2024crosslingualknowledgeeditinglarge} and related studies. As shown in Table\ref{tab:data1}, the models exhibit a consistent trade-off: \textbf{Misalignment between Chinese and English edits.}

This pattern is most evident in llama3-8B with MEMIT at batch size 1000. Chinese editing achieves zh\_socre of 61.71\% with strong zh\_Reliability at 63.21\%, yet en\_reliability collapses to 19.55\%. This indicates that while the model can recall edited facts, it struggles to generate fluent English text and fails to maintain consistency with English knowledge.

Similarly, as shown in Table\ref{tab:data2}, we conducted experiments with MEMIT on llama3-8B across different batch sizes, where the misalignment phenomenon is particularly pronounced. On the trans metric, both monolingual Chinese editing and English editing consistently exhibit relatively low values.Moreover, 
\begin{table*}[th]
\centering
\caption{Model Edit targeting higher-level layers of the models}
\scriptsize
\label{tab:data3}
\renewcommand{\arraystretch}{1.0}
\setlength{\tabcolsep}{3pt}
\begin{tabular}{@{} l l l *{9}{c} @{}}
\toprule
\multirow{2}{*}{Model} & 
\multirow{2}{*}{Methods} & 
\multirow{2}{*}{\makecell{Editing\\Language}} & 
\multirow{2}{*}{\makecell{zh\\score}} & 
\multirow{2}{*}{\makecell{en\\score}} & 
\multicolumn{3}{c}{zh} & \multicolumn{3}{c}{en} & 
\multirow{2}{*}{trans} \\
\cmidrule(lr){6-8} \cmidrule(l){9-11}
 & & & & & 
\makecell{Reliability} & 
\makecell{Generality} & 
\makecell{Locality} & 
\makecell{Reliability} & 
\makecell{Generality} & 
\makecell{Locality} & \\
\midrule
\multirow{2}{*}{qwen2-7b} & 
\multirow{2}{*}{MEMIT} & Chinese & 61.71\% & 41.55\% & 60.16\% & 53.35\% & 71.61\% & 19.04\% & 18.22\% & 87.40\% & 31.65\% \\
 & & English & 49.67\% & 67.31\% & 37.36\% & 36.89\% & 74.75\% & 79.53\% & 44.24\% & 78.16\% & 46.98\% \\
\midrule
\multirow{2}{*}{Llama} & 
\multirow{2}{*}{MEMIT} & Chinese & 66.14\% & 43.83\% & 68.88\% & 60.06\% & 69.47\% & 19.64\% & 18.66\% & 93.20\% & 28.51\% \\
 & & English & 59.72\% & 74.21\% & 44.32\% & 44.36\% & 90.48\% & 85.17\% & 44.41\% & 93.05\% & 52.04\% \\
\bottomrule
\end{tabular}
\end{table*}
Table~\ref{tab:data3} shows that editing different MLP layers 
does not substantially alter this pattern—the cross-lingual misalignment persists 
across all layer choices, confirming it is a fundamental property of the editing 
mechanism rather than an artifact of layer selection.

Critically, at batch size 1000, AlphaEdit and PMET may exhibit catastrophic forgetting, with generation metrics falling below 20\% across multiple language conditions. \textbf{This suggests that certain editing algorithms fundamentally compromise multilingual capabilities when applied at scale, representing a practical limitation of current knowledge editing approaches.}

\subsection{The Linear Effect of Mixed Editing}
\begin{figure*}[t]
    \centering
    \includegraphics[width=0.9\linewidth]{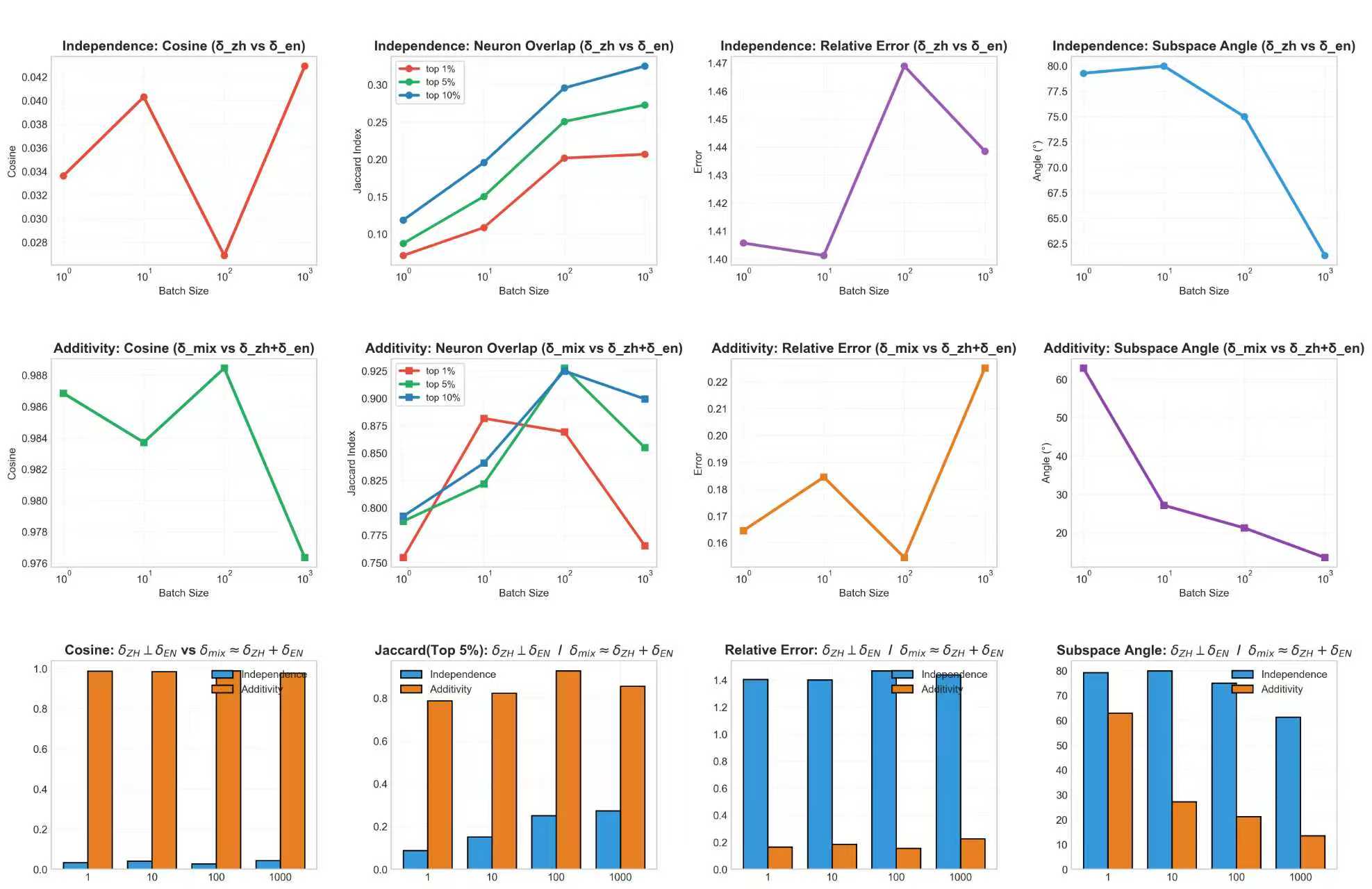}
    \caption{Geometric analysis of language-specific edit vectors across batch sizes. Top: Independence metrics show Chinese and English edits are nearly orthogonal. Bottom: Additivity metrics demonstrate mixed edits closely approximate the linear sum of monolingual edits. Bar charts compare independence (blue) versus additivity (orange), confirming orthogonality between languages and linear composability in mixed editing.}
    \label{fig:analyse}
\end{figure*}
Mixed-language editing results in an approximately linear combination of the monolingual extremes across all configurations. In other words, the Chinese metrics from Chinese editing and the English metrics from English editing roughly sum to the corresponding metrics in mixed editing. This observation further demonstrates that Chinese and English edits are largely independent and additive.

This linearity persists across different batch sizes. As shown in Table\ref{tab:data2}, across batches of 10, 100, and 1000, we observe Chinese editing success rates of 80.00\%, 73.00\%, 82.83\%, and 63.21\%, while the corresponding Chinese editing success rates under mixed editing are 100\%, 94.17\%, 77.20\%, and 63.21\%. The differences are minimal, and similar phenomena are observed in the corresponding English metrics.

\textbf{However, linearity does not imply optimality. Mixed editing may inherit the degraded performance from both monolingual extremes rather than achieving synergistic improvements.} In one scenario, the collapse of either Chinese or English editing directly leads to the collapse of mixed editing. In another scenario, although the Chinese metrics of English editing may exceed those of Chinese editing, mixed editing consistently inherits the lower metrics from Chinese editing. This cumulative interference suggests that practitioners cannot simply hedge linguistic risks—specialized monolingual editing consistently outperforms mixed approaches in their respective target languages.
\section{Analysis and Interpretation}
\subsection{Layer-wise Representation Divergence}
To analyze the differences, independence, and additivity of Chinese-English editing, we extract the mlp\_down\_proj deltas from layers 9, 10, 11, and 12 of Llama3-8B using the MEMIT method. Each layer contains deltas corresponding to Chinese, English, and mixed Chinese-English edits, which are used for subsequent analysis.

To determine which layer's deltas to analyze, we leverage the Bi-ZSRE dataset to construct a language\_vector. We then project and compute the difference between Chinese and English deltas, finding that the language difference intensifies with increasing layer depth. Consequently, we focus our analysis on layer 12 deltas, corresponding to 1, 10, 100, and 1000 edits respectively.More details can be found in Appendix~\ref{app:layer_analysis}.

\subsection{Results \& Analyse of Language-Specific Edit Vectors}

\textbf{Evaluation Metrics}.We evaluate the relation between two edit directions $\Delta_1$ and $\Delta_2$ using four metrics: subspace similarity, relative error, neuron-set overlap, and cosine similarity.

\paragraph{Subspace Angle.}
We apply SVD to each edit matrix $\Delta_1$ and $\Delta_2$, and use the top-$k$ singular vectors to form two subspaces.  
The subspace distance is measured by the principal angles between them.
More details can be found in Appendix~\ref{app:analyse}.

\paragraph{Relative Error.}
To evaluate linear additivity, we compute how well the mixed edit can be reconstructed by the sum of two edits:
\[
\text{RelErr} = 
\frac{\| \Delta_{\text{mix}} - (\Delta_1 + \Delta_2) \|_F}
     {\| \Delta_{\text{mix}} \|_F }.
\]

\paragraph{Neuron Overlap.}
For each edit direction, we select the top-$k$ neurons with the largest absolute changes.  
Let $A$ and $B$ denote these neuron sets for $\Delta_1$ and $\Delta_2$.  
Their overlap is quantified using the Jaccard index:
\[
J = \frac{|A \cap B|}{|A \cup B|}.
\]

\paragraph{Cosine Similarity.}
We flatten both edit matrices and compute their cosine similarity:
\[
\text{CosSim} =
\frac{ \Delta_1 \cdot \Delta_2 }
     {\| \Delta_1 \|_2 \, \| \Delta_2 \|_2 }.
\]
\paragraph{Orthogonality of Language-Specific Edit Vectors.}
As shown in the top row of Figure~\ref{fig:misalignment}, all four independence metrics consistently indicate that $\delta_{\text{zh}}$ and $\delta_{\text{en}}$ form nearly orthogonal directions across batch sizes. The cosine similarity remains extremely small (0.027--0.042), suggesting negligible directional alignment. Neuron overlap, measured by the Jaccard index, stays below 0.32 even at batch size 1000, indicating that the two edits activate largely disjoint neuron sets. The subspace angle stays above $60^\circ$ for all settings, reaching nearly $80^\circ$ for smaller batches---strong geometric evidence that the two edits lie in distinct subspaces. Finally, the relative error remains high ($>1.40$), showing that $\delta_{\text{zh}}$ and $\delta_{\text{en}}$ cannot approximate one another. Taken together, these results demonstrate a stable independence between the two edit directions.
\paragraph{Linear Additivity in Mixed Editing.}
The bottom row of Figure~\ref{fig:misalignment} shows that mixed edits exhibit strong linear additivity. The cosine similarity between $\delta_{\text{mix}}$ and $\delta_{\text{zh}}+\delta_{\text{en}}$ is consistently above 0.976, indicating that their directions are almost identical. The Jaccard index of top-5\% neurons exceeds 0.75 across all batch sizes, suggesting that the neuron set activated by mixed edits is well explained by the union of the monolingual activations. Relative error remains below 0.23 in all cases, showing that the linear sum offers a close reconstruction of the mixed edit. Subspace angles decrease from $63^\circ$ at batch size 1 to $12^\circ$ at batch size 1000, indicating that the mixed edit direction becomes increasingly aligned with the span of the monolingual edit subspaces as more edits are aggregated. These results collectively support that $\delta_{\text{mix}} \approx \delta_{\text{zh}} + \delta_{\text{en}}$ holds robustly across conditions.

\paragraph{Comparative Analysis: Independence versus Additivity.}
The bar charts in Figure~\ref{fig:misalignment} highlight the contrast between independence and additivity. For $\delta_{\text{zh}}$ vs.\ $\delta_{\text{en}}$, independence metrics (blue bars) consistently dominate: cosine similarity is low, neuron overlap is limited, relative error is high, and subspace angles are large. This pattern is stable across batch sizes, indicating that the two monolingual edit directions occupy distinct regions of the parameter space. In contrast, additivity metrics (orange bars) for $\delta_{\text{mix}}$ vs.\ $\delta_{\text{zh}}+\delta_{\text{en}}$ uniformly exhibit the opposite trend---high cosine similarity, substantial neuron overlap, low relative error, and small subspace angles. This reversal shows that although monolingual edits are geometrically independent, their linear combination closely matches the mixed edit direction. The consistency of this behavior across batch sizes suggests that independence and additivity are intrinsic structural properties of how multilingual edits are encoded, rather than artifacts of specific experimental settings.
\section{Conclusion}

We present a culturally grounded Chinese knowledge-editing dataset designed to avoid translation artifacts and to better reflect native factual usage. 
Using this dataset, we evaluate batch editing methods across Chinese and English and identify clear cross-lingual inconsistencies that do not appear in monolingual settings. 
Our mechanistic analysis shows that edits in both Chinese and English exhibit linearity and independence, allowing them to combine additively without interfering with each other.
These findings highlight structural limitations in current editing methods and provide a more realistic basis for assessing multilingual editing reliability.

\section*{Limitations}

This study is limited to Chinese and English, and the extent to which our findings extend to other languages remains an open question. 
Our evaluation centers specifically on batch-editing approaches within the locate-and-edit paradigm, and does not encompass alternative families of editing methods. 
In addition, the mechanistic analysis is confined to a single representative layer of one class of methods, which provides only a partial view of the underlying model dynamics.

\bibliography{custom}
\clearpage
\appendix
\begin{figure}[t]
    \centering
    \includegraphics[width=0.92\linewidth]{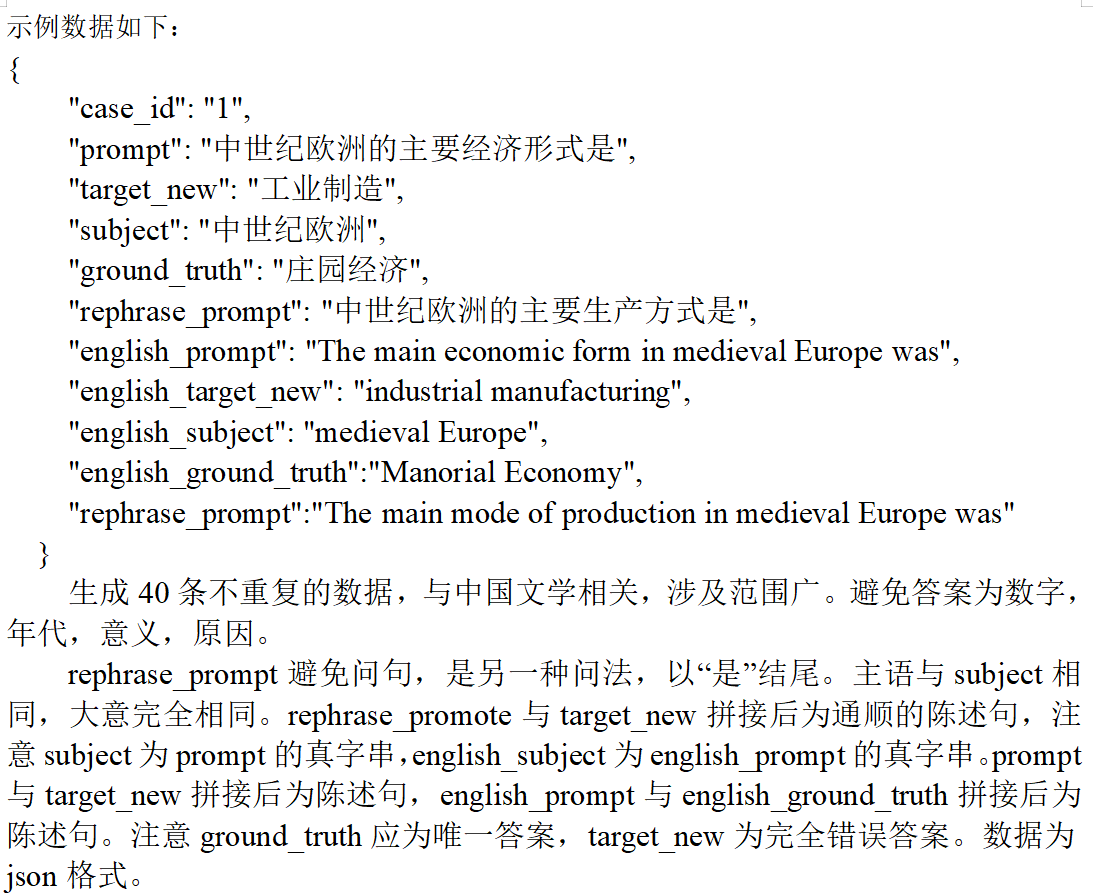}
    \caption{Cross-lingual edit statistics across layers 9--12, including change distributions, neuron comparisons, CDFs, overlap, and clustering structure.}
    \label{fig:prompt}
\end{figure}
\section{Data Collection}
\label{app:data}
\paragraph{(a) Prompt Design.}
We manually categorized the target knowledge into 24 domains, standardized the dataset schema, and specified the syntactic structure for each entry. To avoid low-quality generation, we designed domain-specific prompts. An example is shown in Figure~\ref{fig:prompt}.

\paragraph{(b) Choice of DeepSeek-R1.}
DeepSeek incorporates extensive multilingual data during training, with a particular emphasis on balancing and improving the quality of Chinese–English data. Moreover, DeepSeek enables substantial cost reduction while maintaining high-quality outputs.

\paragraph{(c) Use of ZSRE.}
For the \texttt{loc} and \texttt{loc\_ans} components, we adopt the open-source ZSRE dataset. As a widely used benchmark, ZSRE ensures the reliability and consistency of knowledge localization annotations.

\paragraph{(d) Domain Selection.}
We select domains that are well-represented in both Chinese and English, ensuring high-quality translation, entity coverage, and semantic alignment.

\subsection*{3.2 Post-processing}

\paragraph{(a) Deduplication.}
We applied a Python-based similarity filtering pipeline to remove entries with textual similarity greater than 0.9.

\paragraph{(b) Quality Control.}
To ensure maximal data quality, we conducted manual review. We removed samples with incorrect ground-truth labels, verified that \texttt{target\_new} provides a clearly contrasting incorrect answer, and avoided vague or ambiguous concepts. Reviewers with strong bilingual proficiency eliminated entries with poor translation quality.

\section{Layer Selection Analysis}
\label{app:layer_analysis}

To identify the optimal layer for delta-based cross-lingual analysis, we extract hidden states from 1,000 Chinese–English sentence pairs from the Bi-ZSRE dataset. For each sentence, we use the hidden state of the final token of the subject span and compute layer-specific language vectors across layers 9–12.

\subsection{Language Separation Across Layers}

\begin{figure}[t]
    \centering
    \includegraphics[width=0.92\linewidth]{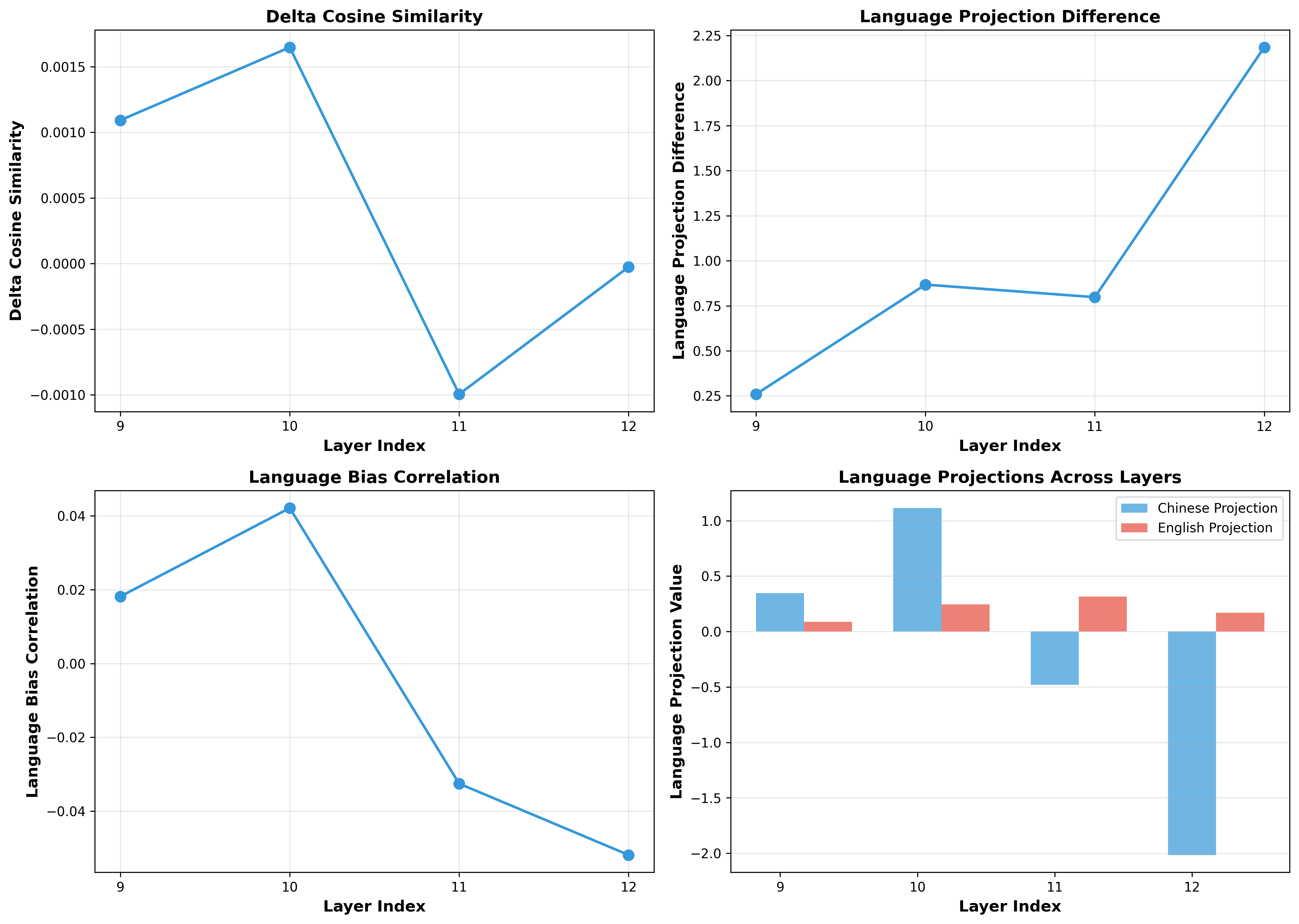}
    \caption{Language-specific patterns across layers 9--12, including delta cosine similarity (top-left), projection difference (top-right), bias correlation (bottom-left), and raw projections (bottom-right).}
    \label{fig:layer_analysis}
\end{figure}

Figure~\ref{fig:layer_analysis} summarizes multiple language-separation metrics across layers.  
The top-left panel shows that delta cosine similarity peaks at layer 10 and drops sharply at layer 11, indicating increased divergence between language-specific deltas.  
The top-right panel shows a monotonic growth in language projection difference from layer 9 to 12, exceeding 2.0 in the deepest layer.  
The bottom-left panel displays language-bias correlation, which peaks at layer 10 before diminishing.  
The bottom-right panel presents raw projection values, where Chinese and English diverge substantially at layer 12 (Chinese near $-2.0$, English near $0.2$).

\subsection{Dimensionality Reduction Visualization}

\begin{figure}[t]
    \centering
    \includegraphics[width=0.92\linewidth]{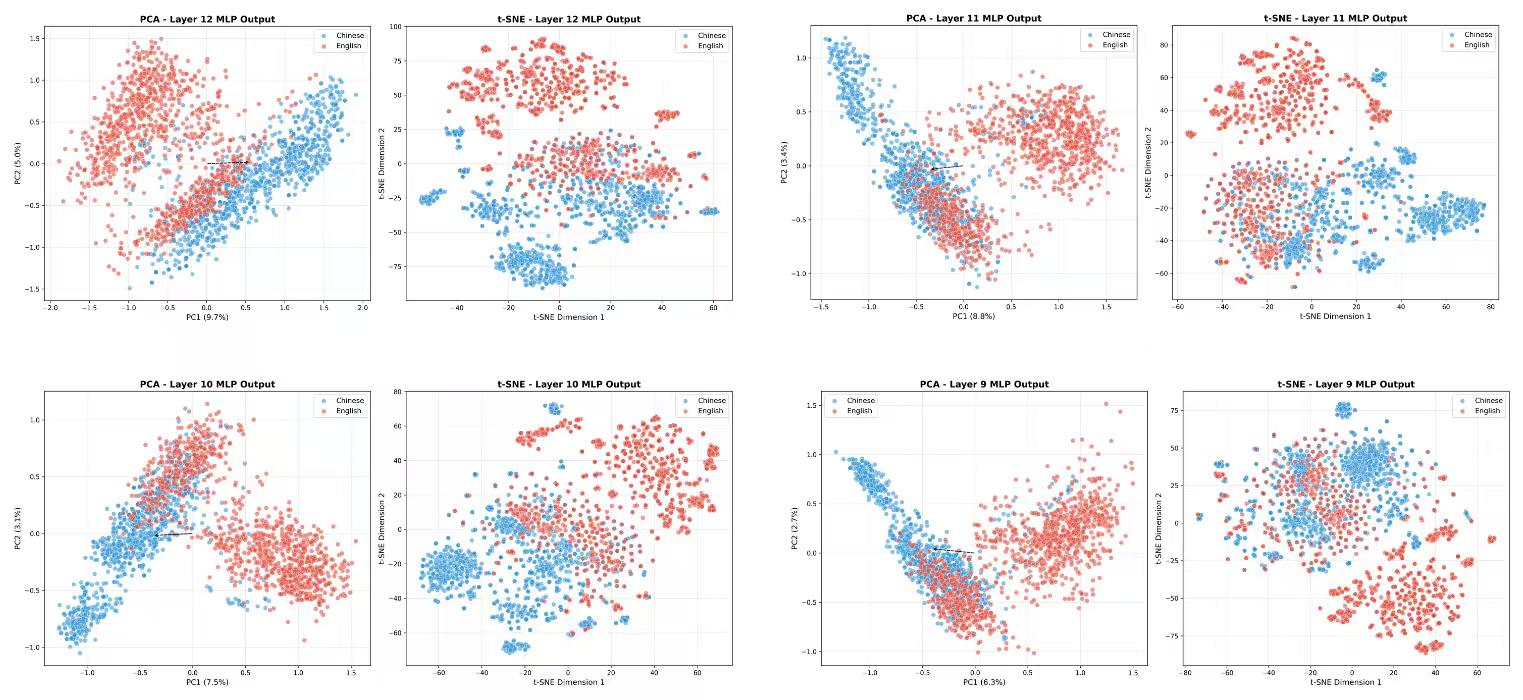}
    \caption{PCA and t-SNE visualization of MLP outputs (layers 9--12). Blue: Chinese; Red: English. Language clusters become increasingly separable in deeper layers.}
    \label{fig:pca_tsne}
\end{figure}

Figure~\ref{fig:pca_tsne} presents PCA and t-SNE visualizations of the MLP outputs at layers 9–12.  
Across all layers, Chinese (blue) and English (red) representations are clearly separable, with the most distinct clustering observed at layer 12.  
These visualizations confirm that language-specific information becomes increasingly structured in deeper layers.

\subsection{Language Bias and Edit Selectivity}

\begin{figure}[t]
    \centering
    \includegraphics[width=0.85\linewidth]{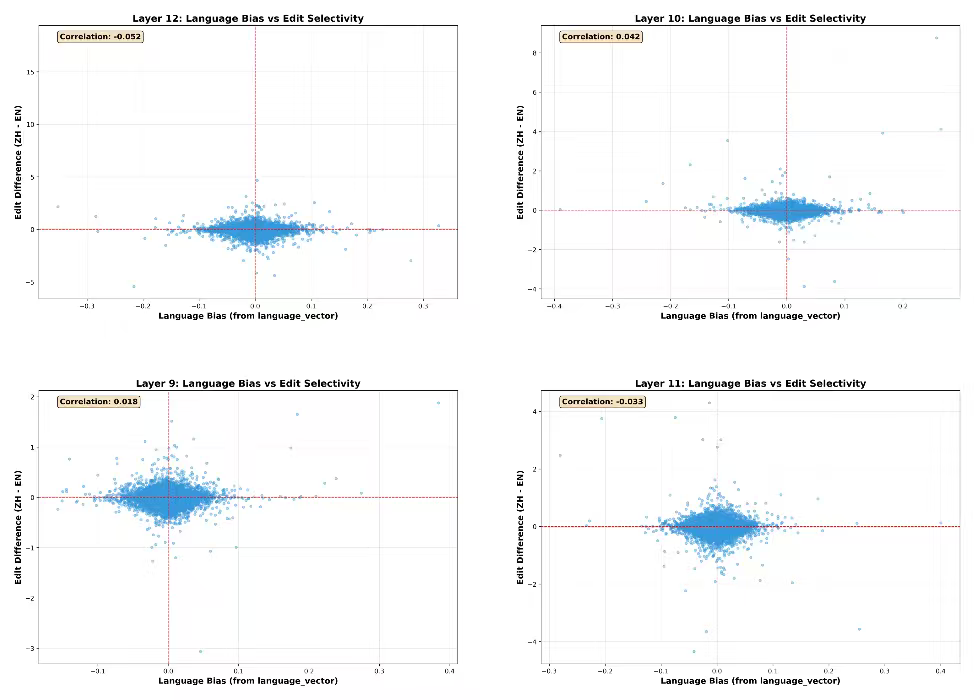}
    \caption{Language bias versus edit selectivity across layers 9--12. Correlation remains near zero, indicating independence.}
    \label{fig:bias_selectivity}
\end{figure}

Figure~\ref{fig:bias_selectivity} analyzes the relationship between language bias and edit selectivity across layers 9–12.  
Correlation coefficients remain near zero (from $-0.092$ to $0.042$), showing that the two factors are statistically independent.  
This indicates that cross-lingual editing challenges cannot be explained purely by language bias.

\subsection{Cross-lingual Edit Distribution}

\begin{figure}[t]
    \centering
    \includegraphics[width=0.92\linewidth]{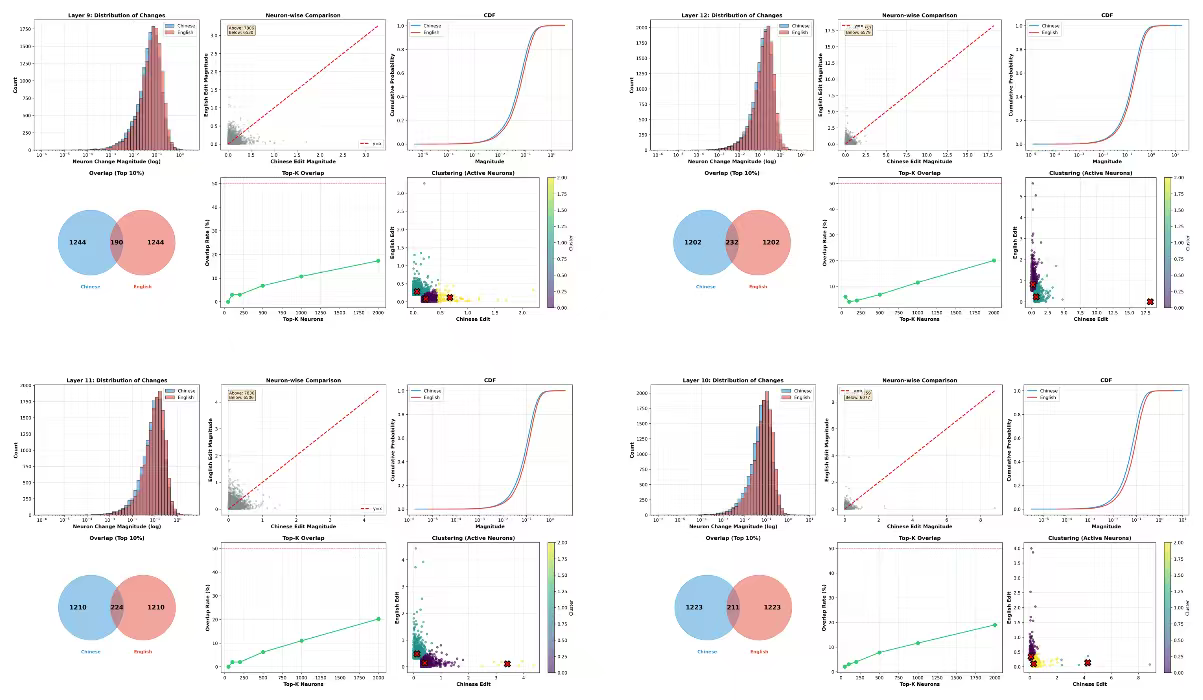}
    \caption{Cross-lingual edit statistics across layers 9--12, including change distributions, neuron comparisons, CDFs, overlap, and clustering structure.}
    \label{fig:edit_distribution}
\end{figure}

Figure~\ref{fig:edit_distribution} provides a comprehensive comparison of cross-lingual edit behavior across layers 9–12.  
The left panels show neuron-wise change distributions, where both languages display similar shapes.  
The middle panels compare actual vs.\ expected activation magnitudes, showing increasing deviation from identity at deeper layers.  
The right panels visualize cumulative distribution functions (CDFs), again indicating broadly similar edit profiles.  
Venn diagrams reveal substantial neuron overlap (around 50\% at layer 12), while clustering plots show dense, structured activation patterns.

\paragraph{Summary.}
Overall, language divergence consistently increases with depth.  
Layer 12 exhibits the strongest separation, the largest projection differences, and the most structured cross-lingual activation patterns.  
Thus, we conduct all subsequent delta-based cross-lingual editing experiments at **layer 12**.
\section{Subspace Angle method}
\label{app:analyse}

Let the edit matrices be $\Delta_1, \Delta_2 \in \mathbb{R}^{m \times n}$.  
We perform singular value decomposition (SVD) on each matrix:
\[
\Delta_1 = U_1 \Sigma_1 V_1^\top, \quad 
\Delta_2 = U_2 \Sigma_2 V_2^\top
\]

The top-$k$ left singular vectors are used to form two subspaces:
\[
\mathcal{S}_1 = \mathrm{span}(U_1^{(k)}), \quad 
\mathcal{S}_2 = \mathrm{span}(U_2^{(k)})
\]

The principal angles $\theta_i$ between the two subspaces are defined as:
\[
\cos \theta_i = \max_{\mathbf{u} \in \mathcal{S}_1} \max_{\mathbf{v} \in \mathcal{S}_2} \mathbf{u}^\top \mathbf{v}, 
\quad i = 1, \dots, k
\]

In our experiments, k is set to 10.
\end{document}